\documentclass[11pt]{article}

\usepackage{algorithmic}

\usepackage{algorithm}
\usepackage{microtype}
\usepackage{graphicx}
\usepackage{subfigure}
\usepackage{booktabs} 
\usepackage{hyperref}
\usepackage{amssymb,amsmath,url,amsthm}
\usepackage{natbib}
\usepackage{wrapfig}
\usepackage{multirow}
\usepackage[dvipsnames]{xcolor}

\usepackage[ margin=1in]{geometry}

\theoremstyle{plain}

\theoremstyle{definition}

\theoremstyle{remark}

\theoremstyle{remark}

\theoremstyle{definition}

\usepackage{bm}

\usepackage{xcolor}

\newenvironment{CompactEnumerate}{
\begin{list}{\arabic{enumi}.}{%
\usecounter{enumi}
\setlength{\leftmargin}{14pt}
\setlength{\itemindent}{1pt}
\setlength{\itemsep}{1pt}
}}
{\end{list}}

\newenvironment{CompactItemize}{
\begin{list}{$\bullet$}{%
\setlength{\leftmargin}{14pt}
\setlength{\itemindent}{0pt}
\setlength{\itemsep}{0pt}
}}
{\end{list}}
\newcommand{\ours}{\textrm{AdaRank}}
\newcommand{\bbm}{\mathfrak{m}}

\newcommand{\bbX}{\bm{X}}

\newcommand{\bbP}{\mathcal{P}}

\begin{document}
\title{\ours{}: Disagreement Based Module Rank Prediction for Low-rank Adaptation}
\date{}
\author{Yihe Dong \\
Google
}

\maketitle

\begin{abstract}
With the rise of language and multimodal models of ever-increasing size, pretraining a general-purpose foundational model and adapting it to downstream tasks has become common practice. To this end, adaptation efficiency can be a critical bottleneck  given the large model sizes, hence efficient finetuning methods such as LoRA have become prevalent. However, LoRA is typically applied with the same rank across all model layers, despite mounting evidence from transfer learning literature that during finetuning, later layers diverge more from pretrained weights. Inspired by the theory and observations around feature learning and module criticality, we develop a simple \textit{model disagreement} based technique to predict the rank of a given module relative to the other modules. 
Empirically, \ours{} generalizes notably better on unseen data than models of uniform ranks with the same number of parameters. Compared to prior work, \ours{} has the unique advantage of leaving the pretraining and adaptation stages completely intact: no need for any additional objectives or regularizers, which can hinder adaptation accuracy and performance. Our code is publicly available at \url{https://github.com/google-research/google-research/tree/master/adaptive_low_rank}.

\end{abstract}

\section{Introduction}
As large language and multimodal models continue increasing in size \citep{gpt, palm}, pretraining a general foundational model and adapting it to specific downstream tasks has become common practice. To this end, adaptation efficiency can be a critical bottleneck given the large model sizes, hence efficient finetuning methods such as LoRA \citep{lora} have become prevalent. However, LoRA is typically applied with the same rank across all model layers, despite transfer learning studies on the similarities between modules before and after adaptation found that later layers deviate more after finetuning \citep{feature_transfer_learning}. Indeed, the similarity between weights before and after finetuning can be an order of magnitude greater in the earliest layers than in the latest layers. This is consistent with the observations that later layers learn finetune-dataset-specific features \citep{yosinski_feature_layers, raghu_feature_layers, zhang_feature_layers}.


This strongly suggests that different layers should use different ranks during adaptation, which has the potential to not only improve expressiveness by allowing some layers to learn more complex features, but also reduce overfitting by restricting the degrees of freedom in the remaining layers. Of course, achieving this requires a mechanism that distributes parameters effectively. Towards this, we propose \ours{}, a simple two-step algorithm for layerwise rank prediction for low rank adaptation.

In summary, our main \textbf{contributions} are:
\begin{CompactItemize}
\item We develop \ours{}, a simple two-step mechanism for determining layerwise ranks for low-rank adaptation, based on model disagreements induced by random module perturbations. This is rooted in principles around module criticality.
\item \ours{} leaves pretraining and adaptation completely intact, without the need for additional objectives for rank prediction that can slow down convergence.
\item Empirically, applying \ours{}-predicted layerwise custom ranks allocates parameters more effectively, and improves model generalizability compared with uniform ranks for all layers with the same total number of parameters. 
\end{CompactItemize}

Indeed, if different layers learn different types of features \citep{feature_transfer_learning, yosinski_feature_layers, raghu_feature_layers}, which inherently vary in their information-theoretic content, then mandating all layers to have the same width can lead to overfitting on certain features and reduced model generalizability.
We hypothesize that \ours{} can improve model generalization and reduce overfitting with its \textbf{more judicious allocation of parameters} across layers. Specifically, more critical modules receive finer-grained updates during adaptation via higher ranks, while less critical modules receive coarser-grained updates.

\begin{figure}
    \centering
    \includegraphics[width=0.4\linewidth]{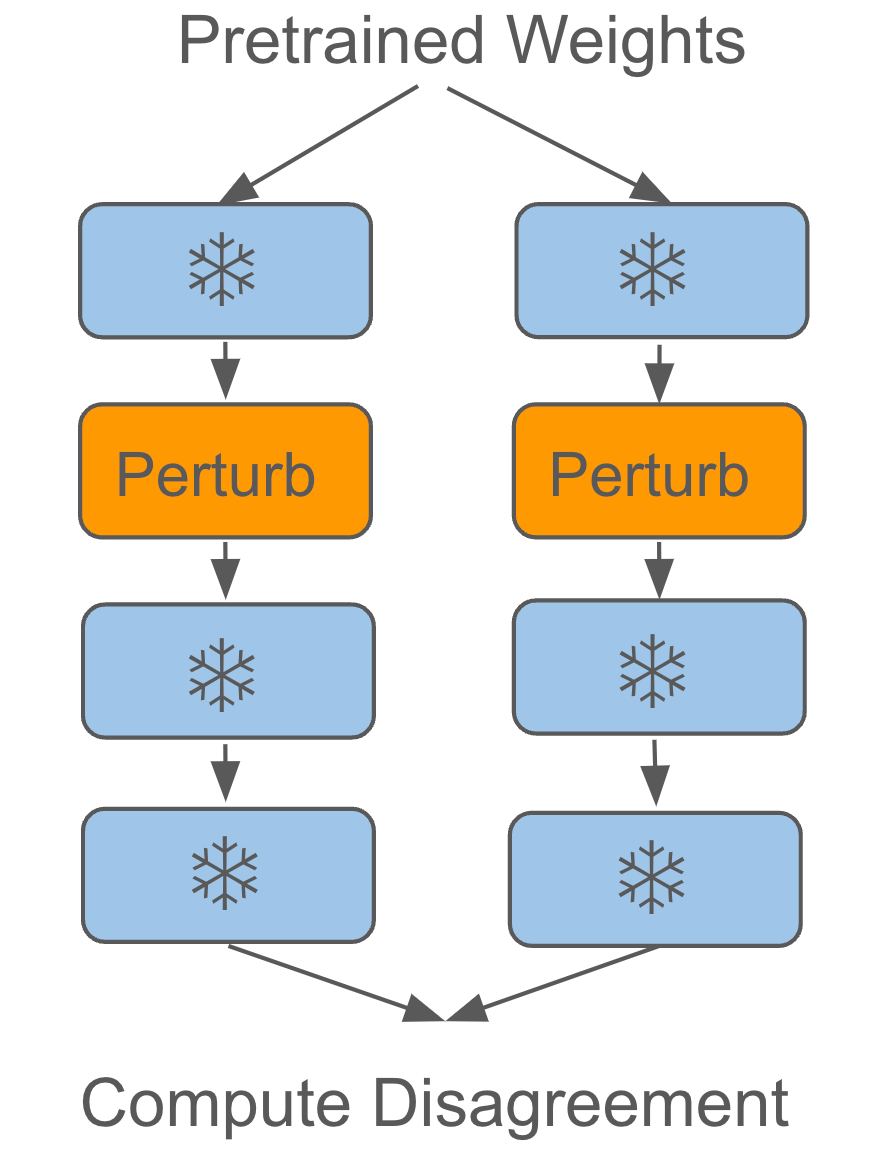}
    \caption{Disagreement-based module relative importance prediction in \ours{}. For a module $\bbm$ (in \textcolor{Orange}{orange}), to approximate its importance score, we 1) add a random noise tensor of the same shape to $\bbm$'s pretrained weights, while keeping all other modules frozen (in \textcolor{CornflowerBlue}{blue}), 2) do this twice, and 3) take the $\ell_1$ difference between model logits produced by these two perturbed model instances. 
    }
    \label{fig:perturbation}
\end{figure}

\paragraph{Notation:}

Module $\bbm$ broadly refers to a subcomponent in the model: 
such as the query, key, or value weights within the attention mechanism, or the dense feedforward module immediately after the attention module.
We denote the pretrained model as $\bbP$, and use $\bbP_{\bbm}$ to denote a perturbed model instance with all weights except the module $\bbm$ and the prediction head frozen, and random noise tensor $\bm{\delta}$ added to $\bbm$. $\bbX$ denotes the input samples used to calculate model output disagreement.


\begin{algorithm}[tb]
   \caption{\ours{} overview.}
   \label{alg:adarank}
\begin{algorithmic}
   \STATE {\bfseries Input:} Pretrained model $\bbP$, list of modules $\{\bbm\}$ across layers to predict ranks on.
   \STATE {\bfseries Input:} Random input sample text $\bbX$.
   \STATE {\bfseries Input:} Desired average rank $r$.
   \STATE {\bfseries Output:} \ours{} predicted \textbf{ranks}.
   \STATE 1) \textit{Module importance prediction}: call Algorithm~\ref{alg:predict-importance} with $\bbP$, $\{\bbm\}$, and $\bbX$ to obtain disagreement-based module importance scores $\bm{d}$.
   \STATE 2) \textit{Scaling module importance to ranks}: call Algorithm~\ref{alg:disagreement-to-rank} with $\bm{d}$ and $r$ to scale $\bm{d}$ to obtain \textbf{ranks}.
\STATE \textbf{Return} \ours{} predicted \textbf{ranks}.
\end{algorithmic}
\end{algorithm}

\section{Disagreement-based module rank prediction}
As outlined in Algorithm~\ref{alg:adarank}, \ours{} uses a strikingly simple two-step process to predict adaptation ranks for a module at each layer: \textbf{1)} calculate relative module importance scores based on perturbation-induced model disagreements, then \textbf{2)} turn these importance scores into ranks. Algorithm~\ref{alg:predict-importance} outlines the first step, and Algorithm~\ref{alg:disagreement-to-rank} the second. \S\ref{sec:motivation} provides further motivation behind \ours{}.

\ours{} has numerous strengths:
\begin{CompactEnumerate}
\item Leaves pretraining and finetuning completely intact. Unlike prior work such as \citep{adalora, sora}, there is no need for additional objectives or regularizers, which can slow down convergence and affect the optimum reached.
\item It is simple to calculate relative module importances and hence predict ranks: simply perturb module weights multiple times and measure the output disagreements. No training is necessary.
\item No labels needed on the adaptation task, since disagreement can be determined using model logits. This increases the applicability to tasks without supervised labels, and makes the predicted ranks unaffected by mislabels. 
\item Empirically, models using \ours{}-predicted ranks generalize well to unseen data, even when using \textit{generic text}, rather than task-specific text, as input during module importance estimation. This means that \ours{} does not require any task-specific processing or training, hence making predicted ranks easily transferable across datasets and tasks.
\end{CompactEnumerate}
This section is organized as follows: \S\ref{sec:disagreement} describes disagreement based module importance scoring, \S\ref{sec:importance-to-rank} describes how \ours{} turns the importance scores to rank predictions, and \S\ref{sec:motivation} discusses motivations and principles behind \ours{}.
Unless specified otherwise, all experiments are done on the Bert base cased model \citep{bert}.


\begin{algorithm}[tb]
   \caption{Disagreement-based module importance prediction.}
   \label{alg:predict-importance}
\begin{algorithmic}
   \STATE {\bfseries Input:} Pretrained model $\bbP$, list of modules $\{\bbm\}$ across layers to predict ranks on.
   \STATE {\bfseries Input:} Random input sample text $\bbX$.
   \STATE {\bfseries Output:} Relative module importances in terms of impact on model output.
   \FOR{module $\bbm$ in $\{\bbm\}$} 
   \STATE Freeze all weights except module $\bbm$.
   \STATE Create two perturbed model instances $\bbP_{\bbm}$ and $\bbP'_{\bbm}$: each created by perturbing $\bbm$ with random noise tensors from $\mathcal{N}(0, \text{std}(\bbm))$.
   \STATE Evaluate $\bbm$'s impact on model output via disagreement in model logits:  $\bm{d}_{\bbm} = |\bbP_{\bbm}(\bbX) - \bbP'_{\bbm}(\bbX)|_1$.
   \ENDFOR
\STATE \textbf{Return} Relative module importance for each $\bbm$, as a vector of disagreement rates $\bm{d}$.
\end{algorithmic}
\end{algorithm}

\subsection{Module importance  based on model disagreement}
\label{sec:disagreement}
Figure~\ref{fig:perturbation} illustrates the disagreement-based module relative importance prediction in \ours{}. For a module $\bbm$, to approximate its importance score, we 1) add a random noise tensor of the same shape to $\bbm$'s pretrained weights, while keeping all other modules frozen, 2) do this twice, and 3) take the $\ell_1$ difference between model logits produced by these two perturbed model instances.

The noise tensor added to each module weight tensor consists of i.i.d. samples from a zero-centered normal distribution, whose standard deviation is the same as the standard deviation of the module $\bbm$ weight tensor. 

More precisely, to create a perturbed model instance $\bbP_{\bbm}$ for a given module $\bbm$, we simply add a Gaussian noise vector $\bm{\delta}$ to $\bbm$'s pretrained weights, where the entries of $\bm{\delta}$ are i.i.d. samples from $\mathcal{N}(0, \text{std}(\bbm))$.
In other words, each module $\hat{\bbm}$ in $\bbP_{\bbm}$ is created by:

\begin{align} %
\hat{\bbm}  = 
\begin{cases}
 \bbm + \bm{\delta} \ \ \ &\text{if } \text{index}(\hat{\bbm}) = \text{index}(\bbm), \\
 \bbm \ \ \  &\text{otherwise.}
\end{cases}    
\end{align}

We then take a sample input, and run inference on two perturbed module instances $\bbP_\bbm$ and $\bbP_{\bbm'}$, and use the $\ell_1$ difference between their output logits as the module importance score for $\bbm$. This process is outlined in Algorithm~\ref{alg:predict-importance}.

Note that the formulation for the noise vector $\bm{\delta}$ deliberately accounts for the varying levels of standard deviations across layers when perturbing the module $\bbm$.

In practice, we calculate the model disagreements five times, and take the average. In a Google Colab with a single CPU, calculating module importance scores fives times for either the query, key, value, or dense module across all 12 layers in the Bert base model takes less than two minutes.
It is interesting future work to study the trade-off between increasing the number of perturbed model instances $\bbP_\bbm$ when determining module importance scores, versus increasing the number of input samples when determining model output disagreement.


Algorithm~\ref{alg:predict-importance} uses text samples $\bbX$ to run inference on perturbed models. 
For simplicity and ease of use, we use ten sentences of generic text as $\bbX$ for all datasets, which means running \ours{} predictions only once for all datasets for a given desired average rank.
\S\ref{sec:custom-text} discusses using task-specific text as model input for determining module relative importance.

\begin{algorithm}[tb]
   \caption{Converting disagreement-based module importance rates into rank prediction.}
   \label{alg:disagreement-to-rank}
\begin{algorithmic}
   \STATE {\bfseries Input:} Module disagreement rate vector $\bm{d}$ for all layers, as computed by Algorithm~\ref{alg:predict-importance}.
   \STATE {\bfseries Input:} Desired average rank $r$.
   \STATE {\bfseries Output:} \ours{} predicted ranks.

   \STATE Compute the average layer disagreement rate: $\overline{\bm{d}}$ = Mean$(\bm{d})$.
   \STATE Compute predicted ranks: \[\textbf{ranks} = \text{Floor}[\bm{d} * (r / \overline{\bm{d}}) ].\]
\STATE \textbf{Return} \ours{} predicted \textbf{ranks}.
\end{algorithmic}
\end{algorithm}

\subsection{From relative importance to absolute ranks}
\label{sec:importance-to-rank}
Algorithm~\ref{alg:predict-importance} predicts relative importance between modules, which needs to be turned into absolute ranks to be used for tuning. 

Algorithm~\ref{alg:disagreement-to-rank} details the conversion from relative module disagreement rate vector $\bm{d}$ to absolute ranks: given the desired average rank $r$ based on resource constraints, we scale $\bm{d}$ to floating points, such that the average rank of the rescaled vector is $r$, then apply the floor function to obtain integral ranks. This produces a set of ranks for modules whose total number of trainable parameters is either the same or just below the number of parameters trained during low-rank adaptation with uniform rank $r$.

There are ways of converting module disagreements into ranks, without an a priori average rank: for instance, we can take the least important module, i.e. the one with the \textit{least} disagreement rate, make its rank 1, and set the ranks for the other layers based on how their disagreement rates compare with the least disagreement rate. We leave this direction to future work.


\subsection{Motivation behind \ours{}}
\label{sec:method-motivation}
This section provides some motivations behind disagreement-based rank prediction, from two points of view: module criticality, and the relation between model agreement and generalization. Section \S\ref{sec:motivation} elaborates further on these motivations.

\paragraph{Module criticality.}  The criticality of a module \citep{moduleCriticalityChatterji2020} refers to how sensitive the overall model output is to changes in that module.
Perturbing a module with random noise and testing the disagreement can be used to assess the criticality of that module. A guiding principle behind \ours{} is that the more critical a module is, the more resources should be allocated towards its finetuning.

Studies have shown that different modules exhibit different levels of criticality, specifically, later layers exhibit higher criticality, i.e. their perturbations affect the output more. This is consistent with the widely known observations that earlier modules learn general features, and later modules learn finer-grained, task-specific features \citep{feature_transfer_learning, yosinski_feature_layers, raghu_feature_layers}.

\paragraph{Model agreement and generalization.} Another motivation behind \ours{} stems from the relation between model agreement and model generalization \citep{kolter, disagreement_generalization}, which found that the smaller the disagreement is between multiple pretrained instances of a model on a test set, the better the model generalizes to that test set. This phenomenon is rooted in the relation between calibration and generalization.

Applying this principle to \ours{}: the smaller the disagreement is between different model instances after perturbing the module, the better the model would be able to generalize, i.e. the features learned by that module during pretraining are robust and generalizable, and hence a lower rank is needed during adaptation.

\begin{table}[t]
\caption{
Statistics on the experimental datasets. These datasets were chosen given their varying domains, sizes, and tasks.
}
\label{tab:data-stats}
\vskip 0.15in
\begin{center}
\begin{small}
\begin{sc}
\resizebox{.7\linewidth}{!}{
\begin{tabular}{lccccr}
\toprule
Dataset & Metric & Train size & Test size & \# classes  \\
\midrule
Trec  & Accuracy & 5,452 & 500 & 6 \\
Yelp  & AUC & 560,000 & 38,000 & 2\\
AG News & Accuracy & 120,000  & 7,600 & 4 \\
\bottomrule
\end{tabular}
}
\end{sc}
\end{small}
\end{center}
\end{table}

\section{Experiments and Results}

This section describes the datasets, experiments, and results on \ours{}. The optimal hyperparameters, e.g. learning rate or batch size, are determined for each experiment with a grid search.
Unless otherwise stated, each result is the average of three runs over different seeds.

\subsection{Datasets}
We test \ours{} on the text classification datasets Trec \citep{trec}, Yelp Reviews \citep{yelp_dataset}, and AG News \citep{ag_news_dataset}. 
Trec \citep{trec} is a question type classification dataset with six classes for questions about topics such as locations, people, numeric information, etc.
The Yelp dataset \citep{yelp_dataset} consists of reviews from the Yelp website, and the target is to classify the sentiment for each review. This binary classification dataset was created by converting one- or two-starred reviews as negative, and three- or four-starred reviews as positive. 
The AG News dataset \citep{ag_news_dataset} consists of news descriptions from more than 2000 news sources, with the goal of classifying each description under categories such as sports or business. Table~\ref{tab:data-stats} contains further statistics on these datasets.


\subsection{Applying \ours{} on individual modules}
\label{section:module_results}

This section discusses applying \ours{} on \textbf{either} the query, key, value, or dense module individually. Here the dense layer refers to the dense MLP layer in the transformer architecture \citep{vaswani} after the self-attention module. 
Applying \ours{} individually gives us a nuanced picture of whether any potential benefits are consistent with respect to each module.

Specifically, we use Algorithm~\ref{alg:predict-importance} followed by Algorithm~\ref{alg:disagreement-to-rank} to predict relative module importances and turn them into ranks. We repeat relative module importance prediction five times, and take the average importance scores as input to Algorithm~\ref{alg:disagreement-to-rank}.

For simplicity and ease of use, we give the \textit{same ten sentences} to Algorithm~\ref{alg:predict-importance} to predict the module importance for all datasets. Appendix~\ref{sec:appendix} lists the sentences used. \S~\ref{sec:custom-text} describes customizing rank prediction using task-specific data.

Following this procedure, here are some examples of the predicted ranks for each module:

\begin{itemize}
  \item Query: $2,4,5,5,6,8,8,9,10,10,11,11$.
  \item Key: $1,2,4,5,7,8,8,10,11,11,11,11$.
  \item Value: $1,3,7,9,9,8,8,8,8,10,10,8$.
  \item Dense: $5,7,7,7,8,9,8,9,8,9,8,6$.
\end{itemize}

\begin{figure}
    \centering
    \includegraphics[width=0.8\linewidth]{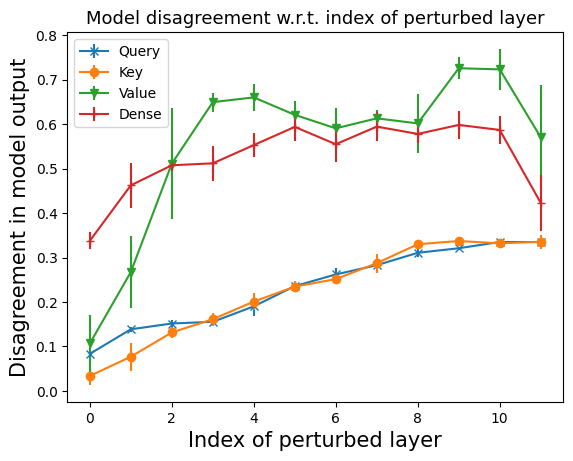}
    \caption{Model disagreement rates when perturbing different modules, perturbed one layer at a time.
    As outlined in Algorithm~\ref{alg:predict-importance}, random noise is added to a given module, while the remaining modules retain their pretrained weights. The model disagreement is calculated as the $\ell_1$ difference in model logits between two perturbed model instances.
    Each data point is the average of five runs. These perturbation rates are used as module importance scores for predicting \ours{} ranks in Algorithm~\ref{alg:disagreement-to-rank}. The upward trends in these model disagreements mirror prior findings that later layers deviate more from pretrained weights during adaptation \citep{moduleCriticalityChatterji2020}.
    Note the varying levels and ranges of disagreement rates for different modules, indicating varying levels of module criticality.
    }
    \label{fig:model_disagreement}
\end{figure}


For both rank 8 (Table~\ref{tab:main-results-separate-modules}) and rank 16 (Table~\ref{tab:main-results-separate-modules-rank-16}), \ours{} consistently outperforms the baseline of applying uniform low ranks to the same modules. This holds for different data regimes with varying dataset sizes. 

Note the average model performance varies when low-rank adaptation is applied to each of the query, key, value, and dense modules. This reveals varying levels of model sensitivity with respect to low-rank adaptation for different modules. One hypothesis is that the higher the model performance is, the less sensitive the model is with respect to lowering the rank from full rank during adaptation.

The \ours{} gains are more prominent for the smaller datasets, especially for the Trec dataset. This is consistent with the hypothesis that \ours{} can improve model generalization and reduce overfitting with its more judicious allocation of parameters, and larger dataset sizes can reglarize the effects of overfitting \citep{dataset_overfitting}.

\begin{table}[t]
\caption{
Performance comparison between uniform ranks and \ours{}-predicted ranks for low-rank adaptation with average rank 8, when adapting either one of the query, key, value, or dense module at a time.
Each result is the average of three runs.
All results are for the Bert model. 
For rank 8, 0.13\% of the non-predictor-head model parameters are trained in each model instance. 
\ours{} is able to more adequately allocate parameters than uniform ranks for improved model generalizability.
}
\label{tab:main-results-separate-modules}
\vskip 0.15in
\begin{center}
\begin{small}
\begin{sc}
\resizebox{.7\linewidth}{!}{
\begin{tabular}{lccccr}
\toprule
Module & Method & Trec $\uparrow$ & Yelp $\uparrow$ & AG News $\uparrow$  \\
\midrule
 \multirow{2}{*}{Query}& \ours{} (ours) & \textbf{95.31} & \textbf{99.44} & \textbf{92.24} \\
 & Uniform & 93.75 & 99.34 & 91.61 \\
 \hline
\multirow{2}{*}{key}& \ours{} & \textbf{96.88} & \underline{99.26} & \textbf{92.21} \\
 & Uniform & 92.97 & 99.20 & 91.79 \\
 \hline
 \multirow{2}{*}{Value}& \ours{} & \textbf{92.19} & \textbf{99.42} & \textbf{92.57} \\
 & Uniform & 91.41 & 99.33 & 91.75 \\
 \hline
\multirow{2}{*}{Dense}& \ours{} & \textbf{92.18} & \underline{99.32} & \textbf{92.15} \\
& Uniform & 87.11 & 99.29 & 91.43 \\
\bottomrule
\end{tabular}
}
\end{sc}
\end{small}
\end{center}
\end{table}

\begin{table}[t]
\caption{
Performance comparison between uniform ranks and \ours{}-predicted ranks for low-rank adaptation with average rank 16, when adapting either one of the query, key, value, or dense module at a time.
Each result is the average of three runs.
All results are for the Bert model. 
For rank 16, 0.26\% of the non-predictor-head model parameters are trained in each model instance. 
Similar to the average-rank-8 experiments, models with \ours{}-predicted ranks demonstrate improved generalizability to the test set. These improvements are more pronounced for the average rank 16 case than for average rank 8.
}
\label{tab:main-results-separate-modules-rank-16}
\vskip 0.15in
\begin{center}
\begin{small}
\begin{sc}
\resizebox{.7\linewidth}{!}{
\begin{tabular}{lccccr}
\toprule
Module & Method & Trec $\uparrow$ & Yelp $\uparrow$ & AG News $\uparrow$  \\
\midrule
 \multirow{2}{*}{Query}& \ours{} (ours) & \textbf{94.53} & \textbf{99.28} & \textbf{92.02} \\
 & Uniform & 92.97 & 99.11 & 91.43 \\
 \hline
\multirow{2}{*}{key}& \ours{} & \textbf{96.48} & \textbf{99.29} & \textbf{92.30} \\
 & Uniform & 91.40 & 99.15 & 91.08 \\
 \hline
 \multirow{2}{*}{Value}& \ours{} & \textbf{95.31} & \textbf{99.59} & \textbf{92.34} \\
 & Uniform & 92.97 & 99.40 & 91.34 \\
 \hline
\multirow{2}{*}{Dense}& \ours{} & \textbf{94.92} & \textbf{99.54} & \textbf{92.31} \\
& Uniform & 93.36 & 99.43 & 91.42 \\
\bottomrule
\end{tabular}
}
\end{sc}
\end{small}
\end{center}
\end{table}

\subsection{Applying \ours{} on all modules simultaneously}
Section \S\ref{section:module_results} demonstrates the consistent improvements seen with \ours{} predicted ranks on individual modules, which leads to the natural question of whether these gains would translate to gains when applying \ours{} to \textbf{all} modules simultaneously.


There are two ways of doing so: 1) simply use the ranks predicted separately for each module, or 2) use the disagreement rates for all modules in conjunction, and feed them together to Algorithm~\ref{alg:disagreement-to-rank} to generate ranks for all modules separately.

We use the second approach, in order to predict ranks for each module in context of the other modules. 
Specifically, we predict ranks by feeding the disagreement rates for all query, key, value, and dense modules \textit{simultaneously} into Algorithm~\ref{alg:disagreement-to-rank}.
An example of such predicted ranks for the Bert base model are:

\begin{itemize}
  \item Query: $1,2,3,3,3,4,5,5,6,6,6,6$. 
  \item Key: $1,1,2,3,4,4,5,5,6,7,6,6$.
  \item Value: $2,5,10,13,13,12,12,12,12,15,15,11$.
  \item Dense: $7,9,10,10,11,12,11,12,12,12,12,8$.
\end{itemize}

Table~\ref{tab:main-results-all-modules-rank-8} and Table~\ref{tab:main-results-all-modules-rank-16} show the performance comparison between uniform ranks and \ours{}-predicted ranks with average ranks 8 and 16, respectively, when adapting \textbf{all} of the query, key, value, and dense modules simultaneously. Adapting all modules simultaneously outperforms adapting any individual module in both the adaptive ranks and the uniform ranks cases, with \ours{}-predicted ranks notably outperforming uniform ranks in all cases for average rank 8, and performing on par or better for average rank 16.

The advantages of \ours{} over uniform ranks appear to saturate, this could be because the advantages of using rank 16 are diminishing over rank 8 on these datasets.
Furthermore, the results for applying \ours{} to all modules simultaneously show that the \ours{} gains when adapting individual query, key, value, or dense modules are not additive.

\begin{table}[t]
\caption{
Performance comparison between uniform ranks and \ours{}-predicted ranks for low-rank adaptation with average rank 8, when adapting \textbf{all} of the query, key, value, and dense modules. Each result is the average of three runs. \ours{}-predicted ranks outperforming uniform ranks in all datasets.
}
\label{tab:main-results-all-modules-rank-8}
\vskip 0.15in
\begin{center}
\begin{small}
\begin{sc}
\resizebox{.6\linewidth}{!}{
\begin{tabular}{lccccr}
\toprule
 Method & Trec $\uparrow$ & Yelp $\uparrow$ & AG News $\uparrow$  \\
\midrule
  \ours{} (ours) & \textbf{97.27} & \underline{99.54} & \textbf{93.23} \\
  Uniform & 96.88 & 99.50 & 92.81 \\
\bottomrule
\end{tabular}
}
\end{sc}
\end{small}
\end{center}
\end{table}

\begin{table}[t]
\caption{
Performance comparison between uniform ranks and \ours{}-predicted ranks for low-rank adaptation with average rank 16, when adapting \textbf{all} of the query, key, value, and dense modules.
The advantages of \ours{} over uniform ranks appear to saturate, this could be because the advantages of using rank 16 are diminishing over rank 8 on these datasets: the average performance across datasets is $96.49$ for rank 16, only a slight improvement over $96.39$ for rank 8. Note that similar to the rank 8 case, adapting all modules with rank 16 outperforms adapting any individual module separately with average rank 16 (Table~\ref{tab:main-results-separate-modules-rank-16}). 
}
\label{tab:main-results-all-modules-rank-16}
\vskip 0.15in
\begin{center}
\begin{small}
\begin{sc}
\resizebox{.55\linewidth}{!}{
\begin{tabular}{lccccr}
\toprule
 Method & Trec $\uparrow$ & Yelp $\uparrow$ & AG News $\uparrow$  \\
\midrule
  \ours{} (ours) & 0.9714 & \underline{0.9965} & \textbf{0.9316} \\
  Uniform & 0.9713 & 0.9961 & 0.9275 \\
\bottomrule
\end{tabular}
}
\end{sc}
\end{small}
\end{center}
\end{table}

\subsection{Rank prediction using in-domain text}
\label{sec:custom-text}

Table~\ref{tab:custom-text} contains results for using ten text samples from the training data, compared to ten generic text samples, as input to Algorithm~\ref{alg:predict-importance} to determine module importances and their ranks.
On average, task-specific text outperformed generic text in model performance, indicating the ranks predicted are better suited for solving the task on average. This advantage does not apply across the board, as the models with generic-text-predicted ranks outperformed those using task-specific text for the query and key modules. This suggests that generic text can function as a proxy for task-specific text in Algorithm~\ref{alg:predict-importance}, further easing its usability.

\begin{table}[t]
\caption{
\textbf{Task-specific text for rank prediction}. This table contains results for using ten text samples from the training data, compared to ten generic text samples, as input to Algorithm~\ref{alg:predict-importance} to determine module importances and their ranks.
On average, task-specific text outperformed generic text in model performance, indicating the ranks predicted are better suited for solving the task on average. This advantage does not apply across the board, as the models with generic-text-predicted ranks outperformed those using task-specific text for the query and key modules. This suggests that generic text can function as a proxy for task-specific text in Algorithm~\ref{alg:predict-importance}, further easing its usability.
}
\label{tab:custom-text}
\vskip 0.15in
\begin{center}
\begin{small}
\begin{sc}
\resizebox{.7\linewidth}{!}{
\begin{tabular}{lcccc|c}
\toprule
 Method & Query & Key & Value & Dense & Average\\
\midrule
  Generic text & \textbf{0.9531} & \textbf{0.9688} & 0.9219 & 0.9218 & 0.9414\\
  Training text & 0.9375 & 0.9453 & \textbf{0.95705} & \textbf{0.9609} & \textbf{0.950187} \\
\bottomrule
\end{tabular}
}
\end{sc}
\end{small}
\end{center}
\end{table}

\subsection{Relative relevance between modules}
For transformer-type models, the set of modules to apply low-rank adaptation on can be either query, key, value, dense feedforward modules, or a combination of these.
Interestingly, perturbing different modules resulted in different levels of disagreement, implying varying levels of importance for different modules. Notably, the value and dense module exhibited greater importance with respect to impact on model output. 



\section{Ablation Studies on rank prediction}

In this section, we check that the improved performance shown by \ours{} is in fact due to the improved suitability of the predicted ranks, leading to better model generalizability, instead being the artifact of the module importance to ranks generation process.

To check this, we uniformly randomly generate a floating point vector in the interval $[0, 1]$. Note the exact range does not matter, as the rank generation process outlined in Algorithm~\ref{alg:disagreement-to-rank} performs a mean normalization. We input this randomly generated module importance vector into Algorithm~\ref{alg:disagreement-to-rank} to predict the corresponding ranks, obtaining ranks such as $12,7,11,2,11,7,9,3,9,1,9,9$. We then compare the model performance against vanilla low-rank adaptation with uniform ranks on Trec. 

As shown in Table~\ref{tab:ablation}, the performance with random ranks performed significantly worse than \ours{}, and generally performed worse than uniform ranks.
Interestingly, for the query module, low-rank adaptation with randomly generated ranks outperforms adaptation with uniform ranks, consistent with the observations in Table~\ref{tab:main-results-separate-modules} that making the query module low rank, versus full rank, during adaptation has the least impact on model generalizability.

\begin{table}[t]
\caption{
Ablation study.
The reported results are the averages of three runs.
}
\label{tab:ablation}
\vskip 0.15in
\begin{center}
\begin{small}
\begin{sc}
\resizebox{.45\linewidth}{!}{
\begin{tabular}{lccccr}
\toprule
Module & Method & Trec $\uparrow$  \\
\midrule
 \multirow{2}{*}{Query}& Random ranks & \textbf{94.53}  \\
 & Uniform & 93.75  \\
 \hline
\multirow{2}{*}{key}& Random ranks & 92.58  \\
 & Uniform & \textbf{92.97}  \\
 \hline
 \multirow{2}{*}{Value}& Random ranks & 85.16  \\
 & Uniform & \textbf{91.41}  \\
 \hline
\multirow{2}{*}{Dense}& Random ranks & 62.89  \\
& Uniform & \textbf{87.11}  \\
\bottomrule
\end{tabular}
}
\end{sc}
\end{small}
\end{center}
\end{table}


\section{Related work}

There has been a plethora of works in recent years focused on efficient adaptation of large language and multimodal models. This includes adapters \citep{adapter, prefix_tuning, compacter, few_shot_finetuning, opendelta}, which have come to refer to a broad family of parameter efficient and modular transfer learning methods from the initial bottleneck-style adapter.
We extend another line of work, LoRA \citep{lora}, which improves the finetuning efficiency of large transformer-based models by applying low-rank approximation to the \textit{update} matrices on top of the pretrained weights.
Recognizing the LoRA's suboptimality of rigidly applying the same rank to all layers, recent works such as \citep{adalora} and \citep{sora} have devised ways to customize the rank for each layer. 
Both AdaLoRA and SoRA leverage the connection between the rank and the singular value decomposition of the weight update matrices, to \textit{learn} layerwise ranks \textit{during adaptation}: AdaLoRA learns to prune the singular values of unimportant updates by adding regularizers and objective functions, and SoRA introduces a learnable gate to gradually reduce the rank during adaptation. 

In contrast, \ours{} leaves pretraining and finetuning completely intact. There is no need for additional objectives or regularizers, which can slow down convergence and affect the training optimum reached. \ours{} does not require any task-specific processing or training, hence making predicted ranks easily transferable across datasets and tasks.

Furthermore, to the best of our knowledge, \ours{} is the first to take the point of view of feature learning and module criticality as a motivation for applying adaptive ranks across layers, which motivated the disagreement-based module importance and rank prediction in \ours{}.

\section{Future work}
This work showed that \ours{} can achieve significant gains in model generalization performance on unseen data, by allocating parameters more effectively across layers. This paves way for a myriad of future directions.

For simplicity, we only used ten generic sentences for the module relative importance prediction in Algorithm~\ref{alg:predict-importance}, it remains to be seen whether using additional input samples can lead to even more pronounced performance gains.
Trade-off between number of trials to average over when generating module relative importances, vs the number of input samples used during each prediction instance.

And although we only observed benefits on average when using in-domain text input for rank prediction, we hypothesize that in-domain text input can lead to more performant rank predictions.
A more thorough investigation in this direction, with more text input over more diverse domains, can shed further insight.

Furthermore, how can we translate the \ours{} performance gains into savings in wall clock timing?
The current experimental setup gives both \ours{} and the uniform rank baseline the same total number of parameters, i.e. same number of FLOPS. However, one can consider turning \ours{} gains into run time savings, by finding the least number of parameters required, when applying \ours{}, to achieve the same performance as uniform ranks. The resulting smaller number of FLOPS can  result in run time savings, as well as savings in the model's memory footprint.

Lastly, while we provided some theory-grounded motivation for \ours{}, it is a very interesting future direction to make the justification rigorous, which can further improve rank prediction.


\section{Acknowledgement}
I am very grateful for the insightful discussions with Ruoxi Wang and Derek Cheng during the development on this work.

\bibliography{references}
\bibliographystyle{plainnat}

\newpage
\appendix

\section{Appendix}

\subsection{Motivation Behind Disagreement Based Rank Allocation}
\label{sec:motivation}

This section elaborates on the motivation behind the rank prediction in \ours{} rooted in the relation between model disagreement and model generalization \citep{kolter}. 

We aim to develop a principled approach, based on the theory that this rank can be predicted using a relation between calibration and generalization. At a high level, \citep{kolter} found that how well a model generalizes to a test set highly correlates with the model's disagreement rate on that test set, where disagreement refers to the difference between multiple pretrained instances of the model, such as using different minibatch orderings during training.

Therefore, if we finetune different instances of a pretrained model, where only a given submodule $\bbm$ is tuned, we can compare the disagreement between these finetuned instances, and determine how relevant that tuned submodule $\bbm$ is towards generalization on the test set. 

In other words, the larger the disagreement is when finetuning a submodule $\bbm$, the harder it is for the model to generalize when training $\bbm$, therefore the more parameter resources, i.e. higher rank, we need to allocate to $\bbm$.

But we don’t want to actually do the finetuning to obtain this disagreement rate. To approximate finetuning, we simply perturb each submodule by a random noise tensor. We hypothesizes that this serves as a sufficiently good approximation, based on NTK theory and linear mode connectivity, and that a well-calibrated pretrained model typically reach a flat basin upon convergence, and finetuning that pretrained checkpoint keeps the weights in that basin. This is to be made rigorous in future studies.


\subsection{Input text used to determine model disagreement}
This section lists the generic input sentences $\bm{X}$ used to determine model disagreements in Algorithm~\ref{alg:predict-importance}:

\begin{CompactItemize}
  \item Here We Go Then, You And I is a 1999 album by Norwegian pop artist Morten Abel. It was Abel's second CD as a solo artist.
  \item  The album went straight to number one on the Norwegian album chart, and sold to double platinum.
  \item Among the singles released from the album were the songs ``Be My Lover" and ``Hard To Stay Awake".
  \item Riccardo Zegna is an Italian jazz musician.
  \item  Rajko Maksimović is a composer, writer, and music pedagogue.
  \item One of the most significant Serbian composers of our time, Maksimović has been and remains active in creating works for different ensembles.
  \item Ceylon spinach is a common name for several plants and may refer to: Basella alba Talinum fruticosum.
  \item A solar eclipse occurs when the Moon passes between Earth and the Sun, thereby totally or partly obscuring the image of the Sun for a viewer on Earth.
  \item A partial solar eclipse occurs in the polar regions of the Earth when the center of the Moon's shadow misses the Earth.
\end{CompactItemize}


\label{sec:appendix}

\end{document}